\documentclass[sigconf]{acmart}
\AtBeginDocument{%
  }

\copyrightyear{2025}
\acmYear{2025}
\setcopyright{cc}
\setcctype{by}
\acmConference[FAccT '25]{The 2025 ACM Conference on Fairness, Accountability, and Transparency}{June 23--26, 2025}{Athens, Greece}
\acmBooktitle{The 2025 ACM Conference on Fairness, Accountability, and Transparency (FAccT '25), June 23--26, 2025, Athens, Greece}\acmDOI{10.1145/3715275.3732031}
\acmISBN{979-8-4007-1482-5/2025/06}

\usepackage{bbm}
\usepackage{tikz}
\usetikzlibrary{positioning}
\theoremstyle{definition}
\newtheorem{prop}{Proposition}

\begin{document}

\title{Aggregating Concepts of Fairness and Accuracy in Prediction Algorithms}

\author{David Kinney}
\orcid{0000-0003-1236-9887}
\affiliation{%
  \institution{Washington University in St.\ Louis}
  \city{St.\ Louis}
  \state{MO}
  \country{USA}
}
\email{kinney@wustl.edu}


\begin{abstract}
  An algorithm that outputs predictions about the state of the world will almost always be designed with the implicit or explicit goal of outputting accurate predictions (i.e., predictions that are likely to be true). In addition, the rise of increasingly powerful predictive algorithms brought about by the recent revolution in artificial intelligence has led to an emphasis on building predictive algorithms that are fair, in the sense that their predictions do not systematically evince bias or bring about harm to certain individuals or groups. This state of affairs presents two conceptual challenges. First, the goals of accuracy and fairness can sometimes be in tension, and there are no obvious normative guidelines for managing the trade-offs between these two desiderata when they arise. Second, there are many distinct ways of measuring both the accuracy and fairness of a predictive algorithm; here too, there are no obvious guidelines on how to aggregate our preferences for predictive algorithms that satisfy disparate measures of fairness and accuracy to various extents. The goal of this paper is to address these challenges by arguing that there are good reasons for using a linear combination of accuracy and fairness metrics to measure the all-things-considered value of a predictive algorithm for agents who care about both accuracy and fairness. My argument depends crucially on a classic result in the preference aggregation literature due to Harsanyi \cite{harsanyi1955cardinal}. After making this formal argument, I apply my result to an analysis of accuracy-fairness trade-offs using the COMPAS dataset compiled by Angwin et al.\ (\cite{angwin2016machine}).
  \end{abstract}




\maketitle

\section{Introduction}
Consider an algorithm that recommends podcasts to users of a digital media service based on data collected, with consent, from those users. Suppose that across all users, there is a strong correlation between a user's gender and their podcast preference:\ in particular, women are much more likely than men to listen to true crime podcasts \cite{beattie2023evaluation}. On a mainstream view, there is some degree of unfairness attached to this algorithm using these correlation data to recommend a true crime podcast to a woman solely because she is a woman, especially if she is a new user with no history of listening to true crime podcasts \cite{yang2020equal}. More serious violations of fairness in predictive systems can occur when a person is harmed in some way (e.g., given a harsher prison sentence) in virtue of an algorithmic prediction based primarily on their possession of a protected characteristic such as race or gender \cite{angwin2016machine}.\par

At the same time, predictive algorithms should strive for accuracy. In the case described above, when an algorithm has little user data to base its predictions on other than basic demographic data, it may be that the best recommendation, from the perspective of expected accuracy, is one that suggests true crime podcasts to women as a blanket policy. Thus, it seems that we are at least sometimes forced to balance a desire for fairness against a desire for accuracy when building predictive systems. There is now a large literature on the need to balance accuracy and fairness in predictive machine learning, with debates over whether the trade-off between accuracy and fairness is inherent or contingent (e.g., \cite{wick2019unlocking,cooper2021emergent,menon2018cost,chen2018my,zhao2022inherent,liu2022accuracy,barlas2021see,vzliobaite2015relation,friedler2021possibility}). See \cite{li2025triangular} for a comprehensive review of this literature. While the balance between accuracy and fairness is, in itself, a difficult balance to strike, the trade-off is made even more difficult by the fact that both accuracy and fairness are highly contested, multi-faceted concepts that can be measured in a multitude of ways, with each measure differing non-trivially from the other. \par

Indeed, an ontology of fairness metrics compiled by Franklin et al.\ contains eighteen different fairness metrics that measure thirty-nine different notions of fairness \cite{franklin2022ontology}. This state of variegation is even more pronounced for measures of accuracy. One desideratum for an accuracy measure that is fairly non-controversial is that it be the expected value of a \textbf{strictly proper scoring rule}. A scoring rule is a function $s$ that provides a measure $s(p,y)$ of the accuracy of a probability distribution $p$ over some set of possible outcomes $Y$ when the actual outcome is $y\in Y$. Such a scoring rule is strictly proper just in case the expected value $\sum_{y\in Y}q(y)s(p,y)$ is maximized if and only if $p=q$. While some scoring rules have become standard in the literature (e.g., the Brier, logarithmic, and spherical scoring rules), there are an \textit{infinite} number of strictly proper scoring rules from which one can choose (\cite{merkle2013choosing}, p.\ 292).\par

Thus, evaluators of predictive algorithms must not only balance fairness against accuracy, but must also balance different concepts of fairness and accuracy that are operationalized by different measures against one another. This paper uses a theoretical result to offer practical guidance for simultaneously managing these trade-offs. Specifically, I show that a classic result due to Harsanyi (\cite{harsanyi1955cardinal}) provides a compelling reason for an agent who cares about both accuracy and fairness to calculate a linear combination of a set of real-valued utility functions, each of which is defined on a measure of accuracy or fairness. To my knowledge, this kind of first-principles argument for the linear aggregation of measures of fairness and accuracy for predictive algorithms has not previously been put forward. While my conclusion does not settle the question of \textit{exactly} how we should balance preferences for various notions of fairness and accuracy when evaluating utility functions, it does tell us the precise mathematical form that such a solution might take, while also clarifying which parts of this balancing act have technical solutions and which parts must be solved through normative argumentation.\par

The remainder of this paper proceeds as follows. In Section \ref{sec:prelim}, I provide the necessary formal preliminaries for the main result. In Section \ref{sec:main}, I argue that measures of accuracy and fairness should be aggregated via linear combination, as per Harsanyi's theorem. In Section \ref{sec:discussion}, I discuss what I take to be the significant implications of this result. In Section \ref{sec:COMPAS}, I use the data set produced by Angwin et al.\ in their audit of the COMPAS algorithm for predictive criminal recidivism (\cite{angwin2016machine}) to illustrate this significance. I conclude in Section \ref{sec:conc}.\par

\section{Formal Preliminaries}\label{sec:prelim}
\subsection{Prediction Algorithms}
Let a \textbf{prediction algorithm} be a function $\mathcal{A}:\mathbf{X}\rightarrow\Delta(Y)$, where $\mathbf{X}$ is a set of \textbf{feature vectors} $\mathbf{x}$ that circumscribes the class of valid inputs to the algorithm, while $\Delta(Y)$ is the set of all probability distributions over a $\sigma$-algebra on an outcome space $Y$. We will suppose that predictions are generated by the algorithm using the following steps:
\begin{enumerate}
    \item We input a feature vector $\mathbf{x}\in\mathbf{X}$.

    \item The algorithm generates a distribution $\mathcal{A}(\mathbf{x})\in \Delta(Y)$.

    \item We sample a value $y$ from the distribution $\mathcal{A}(\mathbf{x})$.
\end{enumerate}
Let $\mathbf{X}^{\prime}\subseteq\mathbf{X}$ be the finite set of inputs that are actually given to the algorithm, and let $Y^{\prime}\subseteq Y$ be the finite set of outputs that are actually produced by the algorithm. This means that we can define a set of input-output pairs $\mathcal{D}=\{(\mathbf{x}_{1},y^{\prime}_{1}),\dots,(\mathbf{x}^{\prime}_{N},y^{\prime}_{N})\}$, where each $y^{\prime}_{i}\sim\mathcal{A}(\mathbf{x}^{\prime}_{i})$. Note that this definition is consistent with a prediction algorithm being entirely deterministic; determinism just requires that for every $\mathbf{x}\in\mathbf{X}$, $\mathcal{A}(\mathbf{x})$ assigns maximal probability to one element of $Y$ and minimal probability to all other elements.\par

\subsection{Accuracy Measures}
To measure the accuracy of a prediction algorithm $\mathcal{A}$ on a finite set $\mathbf{X}^{\prime}\subseteq\mathbf{X}$ of observed feature vectors (where $\mathbf{X}$ is the domain of $\mathcal{A}$), we must first define a \textbf{ground truth function} $F$ on $\mathbf{X}^{\prime}$. For any $\mathbf{x}^{\prime}\in\mathbf{X}$, we interpret $F(\mathbf{x}^{\prime})$ as the actual outcome of some process in the world (specifically, the process whose outcome we are trying to predict) that takes as input the feature vector $\mathbf{x}^{\prime}$. In general, when using the algorithm $\mathcal{A}$ to predict the output of this process when the input is $\mathbf{x}^{\prime}$, we are successful to the extent that $y^{\prime}\sim\mathcal{A}(\mathbf{x}^{\prime})$ is close, in some sense, to $F(\mathbf{x}^{\prime})$. In other words, the ground truth function $F$ provides the \textit{standard of accuracy} according to which a learning algorithm $\mathcal{A}$ is assessed.\par

As mentioned above, we will presuppose here that accuracy can be measured using the expected value of a \textbf{strictly proper scoring rule}. Though we sketched a mathematical definition of such a scoring rule above, here we define it for the specific context of evaluating a prediction algorithm. For any set of outcomes $Y$, a \textbf{scoring rule} is a function $s:\Delta(Y)\times Y\rightarrow \mathbbm{R}$. We interpret $s(\mathcal{A}(\mathbf{x}^{\prime}),F(\mathbf{x}^{\prime}))$ as a measure of the accuracy of the prediction made by the algorithm $\mathcal{A}$ when given the input $\mathbf{x}^{\prime}$, if the ground-truth output for input $\mathbf{x}^{\prime}$ is input is $F(\mathbf{x}^{\prime})\in Y$. As we note above, such a scoring rule is strictly proper just in case, for all $\mathbf{x}^{\prime}$, $\sum_{y\in Y}q(y)s(\mathcal{A}(\mathbf{x}^{\prime}),y)$ is maximized if and only if $q=\mathcal{A}(\mathbf{x}^{\prime})$.\par

One commonly used scoring rule is due to Brier (\cite{brier1950verification}). To define this scoring rule, let $\mathbbm{1}_{F(\mathbf{x}^{\prime})}:Y\rightarrow\{0,1\}$ be a function such that $\mathbbm{1}_{F(\mathbf{x}^{\prime})}(y)=1$ if $y=F(\mathbf{x}^{\prime})$, with $\mathbbm{1}_{F(\mathbf{x}^{\prime})}(y)=0$ otherwise. For an algorithm $\mathcal{A}$, ground-truth function $F$ and input $\mathbf{x}^{\prime}$, the \textbf{Brier score} of the accuracy of $\mathcal{A}$ is given by the equation:
\begin{equation}\label{eq:brier}
    \textsc{Brier}(\mathcal{A}(\mathbf{x}^{\prime})),F(\mathbf{x}^{\prime}))=-\sum_{y\in Y}\left(\mathcal{A}(\mathbf{x}^{\prime})(y) - \mathbbm{1}_{F(\mathbf{x}^{\prime})}(y)\right)^{2}.
\end{equation}
The larger the value of this Brier score, the more accurate the predictions of the algorithm $\mathcal{A}$ are deemed to be (note that on some presentations, the Brier score is define as an \textit{in}accuracy measure to be \textit{minimized}, in which case the RHS of Equation~\ref{eq:brier} is not negated). For the same inputs, the \textbf{logarithmic score} is given by the equation:
\begin{equation}\label{eq:log}
    \textsc{Log}(\mathcal{A}(\mathbf{x}^{\prime})),F(\mathbf{x}^{\prime}))=\ln\left(\mathcal{A}(\mathbf{x}^{\prime})\left(F(\mathbf{x}^{\prime})\right)\right),
\end{equation}
where $\mathcal{A}(\mathbf{x}^{\prime})\left(F(\mathbf{x}^{\prime})\right)$ is the probability assigned the ground-truth output $F(\mathbf{x}^{\prime})$ by the distribution $\mathcal{A}(\mathbf{x}^{\prime})$. Finally, the \textbf{spherical score} is given by the equation:
\begin{equation}\label{eq:spherical}
    \textsc{Sphere}(\mathcal{A}(\mathbf{x}^{\prime})),F(\mathbf{x}^{\prime}))=\frac{\mathcal{A}(\mathbf{x}^{\prime})\left(F(\mathbf{x}^{\prime})\right)}{\sqrt{\sum_{y\in Y}\mathcal{A}(\mathbf{x}^{\prime})(y)}}.
\end{equation}
While all three of these scoring rules typically differ as to the values they return for different inputs, all three are strictly proper scoring rules.\par

While a scoring rule measures the accuracy of a predictive algorithm for a single input, we typically seek to evaluate the accuracy of a predictive algorithm across all actual inputs $\mathbf{X}^{\prime}$. To do this, let $s$ be any strictly proper scoring rule and let $\mathbbm{P}$ be some probability distribution over $\mathbf{X}^{\prime}$. The \textbf{accuracy} of an algorithm $\mathcal{A}$ on the set of actual inputs $\mathbf{X}^{\prime}$, given a ground-truth function $F$ and a distribution $\mathbbm{P}$ over $\mathbf{X}^{\prime}$ is given by the equation:
\begin{equation}\label{eq:acc}
    \textsc{Acc}_{s}(\mathcal{A},\mathbf{X}^{\prime},F,\mathbbm{P}) = \sum_{x^{\prime}\in \mathbf{X}^{\prime}}\mathbbm{P}(\mathbf{x}^{\prime})s(\mathcal{A}(\mathbf{x}^{\prime}),F(\mathbf{x}^{\prime})). 
\end{equation}
Thus, an accuracy measure is the expected value of a scoring rule across all actual inputs to a predictive algorithm. To ground thinking, we note that in many cases, we will set the distribution $\mathbbm{P}$ to be uniform over actual inputs $\mathbf{X}^{\prime}$; these will be just those cases where we care equally about the algorithm making accurate predictions across all inputs.\par

Note that the definition of an accuracy measure given above assumes that we have access to the full probability distribution over outcomes produced by our prediction algorithm for all inputs in $\mathbf{X}^{\prime}$. In practice, however, we may not be able to access the precise probability distribution over all outcomes produced by a predictive algorithm. In such cases, we will have to make an empirical estimate of this probability distribution based on the stream of predictions $\mathcal{D} = \{(\mathbf{x}_{1},y^{\prime}_{1}),\dots,(\mathbf{x}^{\prime}_{N},y^{\prime}_{N})\}$, where each $y^{\prime}_{i}\sim\mathcal{A}(\mathbf{x}^{\prime}_{i})$. To this end, let an \textbf{estimation function} $\mathcal{E}$ be a function that takes as input the full stream of predictions $\mathcal{D}$ and returns a set of probability distributions $\{\mathcal{A}(\mathbf{x}^{\prime})|\mathbf{x}^{\prime}\in\mathbf{X}^{\prime}\}$. To ground thinking, one can think of $\mathcal{E}$ as a maximum likelihood estimation function that learns a conditional probability distribution over outputs for each input that best matches the observed data. The \textbf{estimated accuracy} of $\mathcal{A}$ on a stream of predictions $\mathcal{D}$, given a ground-truth function $F$, distribution $\mathbbm{P}$ over $\mathbf{X}^{\prime}$, and estimation function $\mathcal{E}$ is given by the equation:
\begin{equation}\label{eq:estacc}
    \textsc{EstAcc}_{s}(\mathcal{A},\mathcal{D},F,\mathbbm{P},\mathcal{E}) = \sum_{x^{\prime}\in \mathbf{X}^{\prime}}\mathbbm{P}(\mathbf{x}^{\prime})s(\hat{\mathcal{A}}_{\mathcal{E}}(\mathbf{x}^{\prime})),F(\mathbf{x}^{\prime})), 
\end{equation}
where $\hat{\mathcal{A}}_{\mathcal{E}}(\mathbf{x}^{\prime}))$ is the estimation of the distribution $\mathcal{A}(\mathbf{x}^{\prime})$ produced by the estimation function $\mathcal{E}$; this estimation is based on the stream of predictions $\mathcal{D}$. In what follows, when discussing accuracy measures, I will assume that we are always working with measures of \textit{estimated} accuracy, so as to align my results with actual empirical practice when measuring the accuracy and fairness of predictive algorithms.\par

\subsection{Fairness Measures}
Franklin et al.\ (\cite{franklin2022ontology}) distinguish between \textbf{fairness notions}, or ``definitions of fairness that a model [i.e., a predictive algorithm] can satisfy or fail to satisfy'' and \textbf{fairness metrics}, or ``mathematical formulae for measuring how close a given model is to satisfying some fairness notion'' (p.\ 256). For example, a binary prediction algorithm that assigns people to a positive class or a negative class (e.g., the positive class may be inmates predicted not to re-offend and the negative class may be inmates predicted to re-offend) satisfies the fairness notion of \textbf{equal opportunity} with respect to some grouping of inputs if and only if all groups have the same \textbf{false negative rate}. The false negative rate for a group is the probability that a member of that group is assigned to the positive class by the ground truth function is assigned to the negative class by the algorithm (\cite{verma2018fairness}). There are many possible ways to measure the degree to which an algorithm $\mathcal{A}$ whose actual inputs are vectors $\mathbf{X}^{\prime}$, where $\mathbf{X}^{\prime}$ is the union of feature vectors collected from groups $\mathbf{X}^{\prime}_{1}\subset\mathbf{X}^{\prime}$ and $\mathbf{X}^{\prime}_{2}\subset\mathbf{X}^{\prime}$, satisfies equal opportunity with respect to the groups $\mathbf{X}^{\prime}_{1}$ and $\mathbf{X}^{\prime}_{2}$. To explore one possibility, let $\mathbf{X}^{\prime}_{1,+}$ be the elements of $\mathbf{X}^{\prime}_{1}$ assigned to the positive class by the ground-truth function, and let $\mathbf{X}^{\prime}_{2,+}$ be the elements of $\mathbf{X}^{\prime}_{2}$ assigned to the positive class by the ground-truth function. Then, compute the following difference:
\begin{multline}\label{eq:eqopp}
    \textsc{EqOpp} = -\Bigg| \left[ \sum_{x^{\prime}_{1} \in \mathbf{X}^{\prime}_{1,+}} 
    \frac{\mathbbm{P}(x^{\prime}_{1})}{\mathbbm{P}(X^{\prime}_{1,+})}
    \hat{\mathcal{A}}_{\mathcal{E}}(x^{\prime}_{1})(0) \right] \\
    - \left[ \sum_{x^{\prime}_{2} \in \mathbf{X}^{\prime}_{2,+}} 
    \frac{\mathbbm{P}(x^{\prime}_{2})}{\mathbbm{P}(X^{\prime}_{2,+})}
    \hat{\mathcal{A}}_{\mathcal{E}}(x^{\prime}_{2})(0) \right] \Bigg|,
\end{multline}
where $\hat{\mathcal{A}}_{\mathcal{E}}(x^{\prime})(0)$ is the probability that the algorithm assigns a feature vector $\mathbf{x}^{\prime}$ to the negative class, as estimated by the estimation function $\mathcal{E}$ applied to the data stream $\mathcal{D}$, and $\mathbbm{P}$ is a probability distribution over $\mathbf{X}^{\prime}$. The probabilities $\mathbbm{P}(X^{\prime}_{1,+})$ and $\mathbbm{P}(X^{\prime}_{2,+})$ denote the sum of the probabilities of all input vectors in $X^{\prime}_{1,+}$ and $X^{\prime}_{2,+}$, respectively. Thus, the value $\textsc{EqOpp}$ is the negation of the absolute difference between the average probability of the algorithm returning a false negative for a given member of each group. The greater this value, the greater the extent to which the algorithm is deemed to satisfy equal opportunity.\par

Such a metric likely has some desirable properties and some undesirable properties for an empirical measure of the extent to which a prediction algorithm satisfies equal opportunity. This demonstrates the need for a method that aggregates multiple fairness measures:\ we may care about optimizing different measures of equal opportunity to different degrees. We will likely also care about satisfying other fairness notions beyond equal opportunity, and so will need a way of aggregating fairness metrics that target a variety of fairness notions, as well as metrics that measure different notions of accuracy, into a single, all-things-considered measure of both the accuracy and fairness of a predictive algorithm.\par

However, before we do this, we will need to say more about the general form of a fairness measure. Taking the example of \textsc{EqOpp} as a starting point, note that to calculate this value, we needed all of the same inputs as we did to calculate accuracy, except a scoring rule. Specifically, we needed an algorithm $\mathcal{A}$, a data stream $\mathcal{D}$ (to apply to the estimation function), a ground-truth function $F$ (to define the groups $X^{\prime}_{1,+}$ and $X^{\prime}_{2,+}$), a probability distribution $\mathbbm{P}$ over inputs, and an estimation function $\mathcal{E}$. In addition, a fairness measure requires as a further argument a set of \textbf{instructions}, or further pieces of information needed to calculate the measure. For example, to calculate \textsc{EqOpp}, we needed to know how to divide the set of input feature vectors $\mathbf{X}^{\prime}$ into the groups $\mathbf{X}^{\prime}_{1}$ and $\mathbf{X}^{\prime}_{2}$.\par

Other fairness measures require more complex instructions. Consider \textbf{causal measures of fairness}, as proposed by, among others, Nabi and Shpitser (\cite{nabi2018fair}) and Plecko and Bareinboim (\cite{plevcko2024causal}). To compute a measure of causal fairness, we need to represent the different possible values for each component of each feature vectors as random variables, and then define a directed acyclic graph representing the causal relationships between those variables. From there, we can compute a measure of fairness that explicitly accounts for causal considerations. This directed acyclic graph can, in principle, be written as a bit string that is input as an argument to a function measuring the causal fairness of a predictive algorithm, with different bit strings yielding different outputs.\par

Generalizing from this, we are now able to give a generic definition of a fairness measure. A fairness measure takes as input a prediction algorithm $\mathcal{A}$, a data stream $\mathcal{D}$ consisting of actual inputs to the algorithm and outputs from the algorithm, a ground-truth function $F$, a probability distribution $\mathbbm{P}$, an estimation function $\mathcal{E}$, and instructions $\mathcal{I}$, and returns a real number measuring the fairness of the algorithm. Occasionally, some inputs to the algorithm will be inert:\ while we needed the ground-truth function $F$ to calculate \textsc{EqOpp}, it may not be needed for all measures of fairness, which can, in principle, be completely separated from considerations of accuracy. Moreover, the role of the instructions $\mathcal{I}$ in defining a measure of fairness is highly variegated. In the case of \textsc{EqOpp}, the instructions served to identify two classes of inputs with respect to which equal opportunity was to be compared. In the case of causal measures of fairness, the instructions might specific a directed acyclic graph defined on random variables, a graph which is then used to calculate interventional probabilities. In recognition of this variegation, we stipulate only that instructions must be a bit string (i.e., a code readable by a Turing machine) that specifies the additional inputs beyond those specifically listed that are needed to calculate a fairness measure.\par 

As I have noted throughout this section, different measures of accuracy and fairness have different features that may be desirable or undesirable to some degree or another. In light of this, we must consider how someone who cares, to varying degrees, about different notions of fairness and accuracy should decide on an all-things-considered measure of the quality of a prediction algorithm from the perspective of both fairness and accuracy. In the next section, I will argue for a particular answer to this question:\ such a measure must be a linear combination of utility functions defined on the outputs of different measures of fairness and accuracy.\par 

\section{Aggregating Accuracy and Fairness}\label{sec:main}
\subsection{Utility Functions on the Outputs of Fairness and Accuracy Measures}
It follows from the general definitions of accuracy and fairness given above that in both cases, the range of such a measure will always be the set of real number or some subset thereof. Thus, for any measure $M$ of accuracy or fairness, we can define a utility function $u_{M}:\mathbbm{R}\rightarrow\mathbbm{R}$. This utility function encodes an agent's preferences over different values for measures of fairness and accuracy. For any two possible values $m_{i}$ and $m_{j}$ of a fairness measure $M$, if $u_{M}(m_{i})\geq u_{M}(m_{j})$, then the agent in question thinks that it is at least as good for the measure $M$ to take the value $m_{i}$ as it is for $M$ to take the value $m_{j}$. I have assumed throughout that all measures of accuracy and fairness are designed so that the more accurate or fair an algorithm is, the greater the value of the measure. Since I am concerned here with the utility functions of agents who want to use algorithms that are both accurate and fair, I will assume throughout that $u_{M}(m_{i})\geq u_{M}(m_{j})$ if and only if $m_{i}\geq m_{j}$.\par

At the same time, I wish to allow for significant variance in the form of agents' utility functions over measures of accuracy or fairness. So agents may care very little about improvements in accuracy or fairness above a certain threshold; such agents' preferences over values for a given measure might best be represented by a logarithmic function. By contrast, other agents' utility functions might increase linearly with the values of the accuracy or fairness measure on which they are defined. Defining utility functions on the values of accuracy and fairness metrics allows us to account for these differences in the strength of a given agent's preferences for predictive algorithms that satisfy different measures of accuracy or fairness to different degrees.\par

\subsection{Aggregating Utility Functions}
Suppose that an agent would like a prediction algorithm $\mathcal{A}$ to receive the highest possible scores according to a set of accuracy measures $\textbf{\textsc{EstAcc}} = \{\textsc{EstAcc}_{1},\dots,\textsc{EstAcc}_{Q}\}$ and a set of estimated fairness measure $\textbf{\textsc{EstFair}} = \{\textsc{EstFair}_{1},\dots,\textsc{EstFair}_{T}\}$. Each accuracy measure and fairness measure takes as input the same algorithm $\mathcal{A}$, observed data stream $\mathcal{D}$, ground truth function $F$, probability distribution over inputs $\mathbbm{P}$, and estimation function $\mathcal{E}$. We assume further that all fairness measures receive the same set of instructions $\mathcal{I}$, although only some portions of those instructions may be relevant for a given fairness measure. For each measure $M\in\textbf{\textsc{EstAcc}}\cup\textbf{\textsc{EstFair}}$, we define a utility function $u_{M}$ on the outputs of that function; let \textbf{\textsc{Util}} be the resulting set of utility functions. Now consider a vector $\mathbf{v}\in\mathbbm{R}^{Q+T}$, with each component corresponding to the output of a different function in $\textbf{\textsc{EstAcc}}\cup\textbf{\textsc{EstFair}}$. Our goal is to define a function \textsc{Overall} that takes as input an algorithm $\mathcal{A}$, the vector $\mathbf{v}$, the agent's set of utility functions \textbf{\textsc{Util}}, the observed data stream $\mathcal{D}$, the ground truth function $F$, the probability distribution over inputs $\mathbbm{P}$, the estimation function $\mathcal{E}$, and the instructions $\mathcal{I}$ and returns a real number measuring the all-things-considered value of using the algorithm $\mathcal{A}$ to make predictions for an agent who cares about just those concepts of accuracy and fairness encoded by the measures in $\textbf{\textsc{EstAcc}}\cup\textbf{\textsc{EstFair}}$.\par

Let us stipulate that whatever form that function \textsc{Overall} takes, it must satisfy a crucial desideratum. To state these desideratum, we must first state some preliminary definitions. For each utility function $u_{M}\in\textbf{\textsc{Util}}$, define a utility function $\bar{u}_{M}:\mathbbm{R}^{Q+T}\rightarrow\mathbbm{R}$ such that for any $\mathbf{v}=[v_{1},\dots,v_{Q+T}]$ representing the values of the various measures in $\textbf{\textsc{EstAcc}}\cup\textbf{\textsc{EstFair}}$, $\bar{u}_{M}(\mathbf{v})=u_{M}(v_{M})$, where $v_{M}$ is the component of $\mathbf{v}$ corresponding to measure $M$. In other words, $\bar{u}_{M}$ defines a utility function over vectors representing the values of accuracy and fairness measures for an agent who only cares about the value of measure $M$. Let $\Delta(\mathbbm{R}^{Q+T})$ be the set of probability distributions on a $\sigma$-algebra defined on $\mathbbm{R}^{Q+T}$. Finally, for each measure $M$ in $\textbf{\textsc{EstAcc}}\cup\textbf{\textsc{EstFair}}$, let $\succeq_{M}$ be a preference relation over distributions in $\Delta(\mathbbm{R}^{Q+T})$ such that:
\begin{enumerate}
    \item For any $\mu,\nu\in\Delta(\mathbbm{R}^{Q+T})$, $\mu\succeq_{M}\nu$ iff $\mathbbm{E}_{\mu}(\bar{u}_{M})\geq\mathbbm{E}_{\nu}(\bar{u}_{M})$, where $\mathbbm{E}_{p}(\bar{u}_{M})$ is the expected value of the function $\bar{u}_{M}$ according to the probability distribution $p$.

    \item $\mu\sim_{M}\nu$ iff $\mu\succeq_{M}\nu$ and $\nu\succeq_{M}\mu$.
\end{enumerate}
Finally, let $\succeq_{\textsc{Overall}}$ be a preference relation over $\Delta(\mathbbm{R}^{Q+T})$ such that:
\begin{enumerate}
    \item[(3)] For any $\mu,\nu\in\Delta(\mathbbm{R}^{Q+T})$, $\mu\succeq_{\textsc{Overall}}\nu$ iff $\mathbbm{E}_{\mu}(\textsc{Overall})\geq\mathbbm{E}_{\nu}(\textsc{Overall})$.

    \item[(4)] $\mu\sim_{\textsc{Overall}}\nu$ iff $\mu\succeq_{\textsc{Overall}}\nu$ and $\nu\succeq_{\textsc{Overall}}\mu$.
\end{enumerate}
With these definitions in place, we can now state our core desideratum:
\begin{enumerate}
    \item[(\textbf{D})] Any function \textsc{Overall} must be such that, for all $\mu,\nu\in\Delta(\mathbbm{R}^{Q+T})$, if $\mu\sim_{M}\nu$ for all $M$ in $\textbf{\textsc{EstAcc}}\cup\textbf{\textsc{EstFair}}$, then $\mu\sim_{\textsc{Overall}}\nu$.
\end{enumerate}
In other words, \textbf{D} states that if the utility functions representing each of an agent's interests in maximizing each measure in the set of accuracy and fairness measures that the agent cares about all have the same expected value according to two probability distributions, then the utility function representing their overall interest in accuracy fairness should also have the same expected value according to the same two distributions.\par 

I take \textbf{D} to be a highly plausible constraint on the utility function \textsc{Overall}. That said, the constraint is controversial, and must defended here. At its core, \textbf{D} is an \textbf{ex ante Pareto indifference} principle. Suppose that for each measure $M$ in $\textbf{\textsc{EstAcc}}\cup\textbf{\textsc{EstFair}}$, there were an agent $\alpha_{M}$ who myopically cared only whether a predictive algorithm optimized measure $M$. Now suppose that the value of each measure in $\textbf{\textsc{EstAcc}}\cup\textbf{\textsc{EstFair}}$ will be determined by one of two lotteries, where by `lottery' I just mean any probabilistic mechanism for choosing outcomes:\ the first lottery samples values for the measures in $\textbf{\textsc{EstAcc}}\cup\textbf{\textsc{EstFair}}$ from a distribution $\mu$, and the second lottery samples values for the measures $\textbf{\textsc{EstAcc}}\cup\textbf{\textsc{EstFair}}$ from a distribution $\nu$. The desideratum \textbf{D} effectively says that from an \textit{ex ante} perspective (i.e., before we know which values of the measures in $\textbf{\textsc{EstAcc}}\cup\textbf{\textsc{EstFair}}$ will be sampled from $\mu$ or $\nu$), if all such persons are individually indifferent between determining the actual value of each measure by sampling from $\mu$ versus $\nu$, then any representation of the aggregate preferences of the group of such agents must also be indifferent between determining the actual value of each measure by sampling from $\mu$ versus $\nu$. One can, therefore, view \textbf{D} as a minimal commitment of \textbf{ex ante egalitarianism}. If each agent in a group regards two lotteries as having equal value when they do not know how those lotteries will be resolved, and the group is committed to the egalitarian idea that all group members’ interests are equally important, then, collectively, the group should agree that these same two lotteries have equal value.\par

Ex ante egalitarianism has compelling procedural virtues. It commits us, before we know what the outcome of a measurement will be, to a particular procedure for aggregating our preferences as to the accuracy and fairness of an algorithm. This procedure ensures that all interests are given equal consideration, given our realistic but epistemically impoverished position of uncertainty as to the accuracy and fairness of an algorithm. Indeed, it is for this reason, at least partially, that an \textit{ex ante} Pareto indifference principle is widely deployed in social choice contexts involving uncertainty (\cite{harsanyi1955cardinal,arrow1964role,hammond1981ex,mahtani2017ex}). However, it is not without controversy. Fleurbaey and Voorhoeve (\cite{fleurbaey2013decide}) argue that what matters normatively for groups of agents making decisions under conditions of uncertainty is not whether agents agree \textit{ex ante}, but whether they agree \textit{ex post} (i.e., after all uncertainty is resolved). To adopt Fleurbaey and Voorhoeve’s motivating example to my case, imagine an instance where two lotteries over possible numerical values for measures of fairness and accuracy look equally good to all of the myopic agents described above \textit{ex ante}, but only one of those lotteries runs the risk of yielding a very unfair or very inaccurate outcome according to one measure. The desideratum \textbf{D} requires that the group of myopic agents described above regard the two lotteries as equally valuable. But we might have good reason to prefer a lottery that does not run the risk of leaving one myopic agent very badly off.\par

My response to this worry is to concede that Fleurbaey and Voorhoeve’s argument has significant power in the case where the utility functions to be aggregated represent the \textit{actual well-being of real-world agents}. Failing to consider the actual \textit{ex post} outcomes of different agents becomes problematic once one acknowledges that agents are separate persons whose well-being cannot be divided and transferred to different members of a group at different points in time without some member of the group being wronged. However, this same objection does not hold sway in the case where the utility functions in question do not \textit{literally} represent the well-being of agents, but instead represent the interests of a \textit{hypothetical} myopic agent who cares solely about a particular measure of fairness or accuracy. Indeed, if such agents did exist, we would view them as viciously myopic. For this reason, I hold that we can treat these myopic agents as hypothetical posits that serve to establish the appeal of ex ante egalitarianism in motivating the desideratum \textbf{D}, without running afoul of the concerns regarding the separateness of persons raised in the literature on interpersonal preference aggregation.\par

In light of the preceeding argument in favor of \textbf{D} as a constraint on the aggregate utility function \textsc{Overall}, the following proposition establishes that \textsc{Overall} must be a linear combination of the utilities in \textbf{\textsc{Util}}:
\begin{prop}\label{prop:main}
Any function \textsc{Overall} that satisfies \textbf{D} has the form:
\begin{multline*}
    \textsc{Overall}(\mathcal{A},\mathbf{v},\textbf{\textsc{Util}},\mathcal{D},F,\mathbbm{P},\mathcal{E},\mathcal{I})=\alpha \\ + \ \sum_{M\in \textbf{\textsc{EstAcc}}\cup\textbf{\textsc{EstFair}}}w_{M}\bar{u}_{M}(\mathbf{v}),
\end{multline*}
where $\alpha\in\mathbbm{R}$ and $w_{M}\in\mathbbm{R}$ for all measures $M$.
\end{prop}
\noindent
\begin{proof}[Proof Sketch]
    The proof sketch follows the elegant topological proof of Harsanyi's theorem given by Border (\cite{border1985more}). Recall that $\mathbbm{R}^{Q+T}$ is a set of vectors, where each vector represents a set of possible values for the measures in $\textbf{\textsc{EstAcc}}\cup\textbf{\textsc{EstFair}}$. Let $\Sigma$ be a $\sigma$-algebra defined on $\mathbbm{R}^{Q+T}$, i.e., a collection of subsets of $\mathbbm{R}^{Q+T}$ closed under union, intersection, and complement. We stipulate that each $\bar{u}_{M}:\mathbbm{R}^{Q+T}\rightarrow\mathbbm{R}$ is measurable with respect to $\Sigma$, meaning that for each $x\in\mathbbm{R}$, $\bar{u}^{-1}_{M}(x)\in\Sigma$. Let $B$ be the set of all real-valued functions defined on $\mathbbm{R}^{Q+T}$ that are measurable with respect to $\Sigma$. Let $\mathbf{1}$ be the function that returns 1 for every value of $\mathbbm{R}^{Q+T}$. The set $C=\{\mathbf{1}\}\cup\textbf{{\textsc{Util}}}$ is a subset of $B$. The \textbf{span} $S$ of $C$ is the intersection of all subsets of $B$ that contain $C$. Suppose that for a given $(\mathcal{A},\textbf{\textsc{Util}},\mathcal{D},F,\mathbbm{P},\mathcal{E},\mathcal{I})$, there is no set of real numbers $\{\alpha\}\cup\{w_{M}|M\in\textbf{\textsc{EstAcc}}\cup\textbf{\textsc{EstFair}}\}$ defining a function $z(\mathbf{v}) = \textsc{Overall}(\mathcal{A},\mathbf{v},\textbf{\textsc{Util}},\mathcal{D},F,\mathbbm{P},\mathcal{E},\mathcal{I})=\alpha + \sum_{M\in \textbf{\textsc{EstAcc}}\cup\textbf{\textsc{EstFair}}}w_{M}\bar{u}_{M}(\mathbf{v})$, where $\mathbf{v}\in\mathbbm{R}^{Q+T}$, that satisfies $\mathbf{D}$. Since it is well-known that the span of a set of functions contains all its linear combinations, then this would mean that for any $z:\mathbbm{R}^{Q+T}\rightarrow\mathbbm{R}$ that satisfies $\mathbf{D}$, $z\not\in S$. Border shows that this entails the existence of a countably additive signed measure $\tau$ such that the value of the integral $\int_{\mathbbm{R}^{Q+T}}z\text{d}\tau$ must simultaneously be 0 and 1, which is a contradiction. Thus, any function satisfying $\mathbf{D}$ must be in the span $S$ of $C=\{\mathbf{1}\}\cup\textbf{{\textsc{Util}}}$, and therefore must be a linear combination of the functions in $\textbf{{\textsc{Util}}}$.
\end{proof}
One should interpret each $w_{M}$ as a \textbf{weight} assigned to the value of the measure $M$ for the purposes of calculating the overall value using the prediction algorithm $\mathcal{A}$. This proposition is a direct corollary of Harsanyi's 1995 result (\cite{harsanyi1955cardinal}), which shows that for any set of utility functions defined on a common space $\Omega$, any aggregate utility function satisfying an analog of $\textbf{D}$ must be a linear combination of each utility function in that set. Thus, if an agent who cares about both accuracy and fairness, perhaps measured in various ways, wishes to define a measure of the all-things-considered value of using a predictive algorithm, then they should define an aggregate measure of this all-things-considered value that is a linear combination of utility functions representing their preferences over different possible outputs of measures of fairness and accuracy.\par

Before moving to the discussion, it is worth noting that if one replaces \textbf{D} with other plausible constraints on the function $\textsc{Overall}$, then one reaches a different conclusion from that presented in Proposition~\ref{prop:main}. Indeed, Weymark (\cite{weymark1993harsanyi}) shows that if one replaces \textbf{D} with the following \textbf{weak Pareto} principle, then $\textsc{Overall}$ need not be a linear combination of the utility functions in $\textbf{\textsc{Util}}$:
\begin{enumerate}
    \item[(\textbf{WP})] Any function \textsc{Overall} must be such that, for all $\mu,\nu\in\Delta(\mathbbm{R}^{Q+T})$, if $\mu\succ_{M}\nu$ for all $M$ in $\textbf{\textsc{EstAcc}}\cup\textbf{\textsc{EstFair}}$, then $\mu\succ_{\textsc{Overall}}\nu$, where $\mu\succ_{M}\nu$ iff $\mathbbm{E}_{\mu}(\bar{u}_{M})>\mathbbm{E}_{\nu}(\bar{u}_{M})$.
\end{enumerate}
As Weymark points out, \textbf{WP} is logically independent of \textbf{D}, and so we are, logically speaking, free to embrace the former while rejecting the latter. However, while there may be some social choice contexts in which \textbf{WP} should be accepted and \textbf{D} should be rejected, I have argued above that we have good reasons for accepting \textbf{D} when managing trade-offs between concepts of accuracy and fairness. Thus, we can remain committed to the idea that aggregate measures of the fairness and accuracy of a predictive algorithm should have a linear form.\par 

\section{Discussion}\label{sec:discussion}
There is much to unpack about this argument. First, let us consider the separate roles of the utility functions in $\textbf{\textsc{Util}}$ and the role of the set of weights $\{w_{M}|M\in\textbf{\textsc{EstAcc}}\cup\textbf{\textsc{EstFair}}\}$ in generating an all-things-considered judgment of the value of a predictive algorithm for an agent who cares about both accuracy and fairness. The utility function $u_{M}$ (and its extension $\bar{u}_{M}$ to the space $\mathbbm{R}^{Q+T}$) is \textit{myopic}. It encodes solely how the agent evaluates different possible values of the measure $M$ of accuracy or fairness, and does not represent any comparative judgment about how the agent manages the trade-off between different measures of accuracy and fairness. By contrast, the weights $w_{M}$ used to calculate $\textsc{Overall}$ do serve to manage the comparative importance of different notions of accuracy and fairness to an agent's all-things-considered judgment about the value of a predictive algorithm. Holding all other weights constant, if we increase the value of a weight $w_{M}$, then the overall importance of the accuracy or fairness measure $M$ to our judgment of the all-things-considered value of a predictive algorithm will increase, relative to the other measures being considered.\par

What is there to say about how to set the values of the weights $w_{M}$? One fact to note in response to this questions is that nothing that I have said here guarantees that all fairness or accuracy measures can take the same range of values. For example, one measure that contributes to an agent's overall evaluation of a prediction algorithm might be capable of taking any value in the real numbers, while another is defined solely on the unit interval. The same is true for utility functions defined on these ranges of values. In some cases, we may need to set weights strategically so as to ensure that when one utility function takes a maximal or near-maximal value in its domain, the significance of this measurement is not swamped by the values of another utility function simply due to differences in the definition of their ranges.\par

More broadly, however, I deliberately do not take a stand here on how to set the weights used to calculate the precise value of the function $\textsc{Overall}$ for a given algorithm. Decisions about the importance of different notions of fairness and accuracy to our overall evaluation of a predictive algorithm are best arrived at through a deliberative process, both within and between agents, and are likely not going to be derivable from first principles. To suppose otherwise would be to run afoul of warnings from Selbst et al. (\cite{selbst2019fairness}) about the limitations of formal methods for solving contextual problems.\par

Nevertheless, a lack of guidance as to how \textit{exactly} to set the values of the weights assigned to each measure does not render my contribution here vacuous. Rather, I take my application of Harsanyi's theorem to clarify exactly what should and should not be at stake in debates over how to trade considerations of fairness and accuracy against one another when measuring the all-things-considered value of a prediction algorithm. Indeed, we \textit{should} debate about the appropriate values of the weights $w_{M}$ for each measure $M$ we consider, alongside the shape of the utility functions $u_{M}$. These elements of the formal apparatus encode an agent's \textit{values}, and as such are proper subjects for moral and political debate. What should \textit{not} be up for debate, I argue, is whether, once these weights and utility functions are defined, the overall measure of a predictive algorithm's value should be a linear combination of the utility functions. Unless we are willing to make the costly move of rejecting \textbf{D}, it seems clear that the combination of utility functions resulting an all-things-considered measure of a predictive algorithm's value should be linear.\par

On a related note, I wish to stress that we can read very little into the numerical value of the function \textsc{Overall} for a single combination of its arguments. Rather, the function \textsc{Overall} earns its keep primarily as a means of comparing how an agent with a fixed set of weights and utility functions evaluates the overall value of different prediction algorithms, or evaluates the overall value of the same prediction algorithms under different data streams, or ground truth functions, or probability distributions over inputs, or estimation functions, or instructions. That is, it enables \textit{comparative}, rather than absolute, judgments of the value of different means of making predictive inferences for agents who care about fairness and accuracy in a certain way.\par

Note also that I have assumed here that we do not care about interactions between measures of accuracy and fairness. To illustrate, because we assume that the set $\textbf{\textsc{Util}}$ of utility functions to be aggregated by $\textsc{Overall}$ contains only utility functions that are sensitive to a single measure of accuracy or fairness, we are unable to account for cases where a particular measure of fairness is especially important only when some other measure of accuracy or fairness is low or high. One might be concerned about how realistic this assumption is, in light of the fact that we may often care about interaction effects between measures of accuracy and fairness. In response, I need only clarify that this assumption is made solely for ease of exposition and need not hold in general. Indeed, Proposition~\ref{prop:main} will go through for any set $\textbf{\textsc{Util}}$ of real-valued functions defined on the set of $\mathbbm{R}^{Q+T}$ of vectors representing possible values of measures in $\textbf{\textsc{EstAcc}}\cup\textbf{\textsc{EstFair}}$. Such a set $\textbf{\textsc{Util}}$ could, in principle, include utility functions sensitive to interactions between different measures of accuracy and fairness in $\textbf{\textsc{EstAcc}}\cup\textbf{\textsc{EstFair}}$.\par

Finally, by exploring the value \textsc{Overall} under different combinations of weights and utility functions, we can explore how our evaluation of the overall value of a prediction algorithm changes as we adjust what we care about and how we care about it. For instance, if we find that weighting accuracy measures more heavily while decreasing the weight applied to fairness measures increases our evaluation of a predictive algorithm, then we learn that the algorithm will be more appealing to those who care about accuracy than it is to those who care about fairness. Naturally, if weighting fairness measures more heavily while decreasing the weight applied to accuracy measures increases our evaluation of a predictive algorithm, then we will reach the opposite conclusion. These kinds of analyses can become highly nuanced as we allow ourselves to also adjust the utility functions over values of fairness and accuracy measurements (e.g., switching from a logarithmic to a linear utility function over the values of certain measures). In what follows, we will provide a simple illustration of this method of analysis, using the example of the COMPAS data set.\par

\section{The COMPAS Datset}\label{sec:COMPAS}
The COMPAS (Correctional Offender Management Profiling for Alternative Sanctions) algorithm was an algorithm used in a criminal sentencing context to determine a convicted criminal defendant's likelihood of re-offending based on statistical properties of the defendant, such as their previous criminal history. A watershed audit conducted by Angwin et al.\ (\cite{angwin2016machine}) found evidence of significant unfairness against Black defendants. Specifically, Black defendants who were \textit{not} recidivist after two years being were found to be significantly more likely to be classified by COMPAS as at a high risk of re-offending than White patients who were also not recidivist after two years. Here, I use a dataset compiled as part of that audit (specifically, the dataset of algorithmic predictions and outcomes produced by the use of the COMPAS algorithm in Broward County, Florida) to illustrate how the assignments of different weights and utility functions to different measures of fairness and accuracy influences our overall evaluation of a predictive algorithm.\footnote{See \url{https://github.com/davidbkinney/accuracyandfairness} for all data and code used in this analysis.}\par

I begin by translating the predictions of the COMPAS algorithm $\mathcal{A}$ into probabilistic predictions. For each vector of features $\mathbf{x}^{\prime}$ for a defendant, COMPAS outputs a \textbf{decile prediction} in the form of an integer between zero and ten, inclusive. Lower decile predictions indicate a lower likelihood of recidivism. The set of pairs consisting of feature vectors $\mathbf{x}^{\prime}$ and decile predictions amount to a data stream $\mathcal{D}$. In the analyzed data set, there are 7,212 pairs in this data stream. To translate this decile predictions into probabilistic predictions, I use an estimation function $\mathcal{E}$ yielding estimates $\hat{\mathcal{A}}_{\mathcal{E}}$ that are defined as follows:
\footnotesize
\begin{equation}
    \hat{\mathcal{A}_{\mathcal{E}}}(\mathbf{x}^{\prime}) = 
    \begin{cases} 
        0.0001 & \text{if } \texttt{decile\_prediction } \text{for } \mathbf{x}^{\prime} \text{ is } 0 \\ 
        0.9999 & \text{if } \texttt{decile\_prediction } \text{for }  \mathbf{x}^{\prime} \text{ is } 10 \\
        .1(\texttt{decile\_prediction}) & \text{otherwise}
    \end{cases}
\end{equation}
\normalsize
This estimation method ensures that all probabilities of re-offending are non-extreme, while still increasing monotonically with the value of the decile prediction. Next, note that the COMPAS data set produced by Angwin et al.\ effectively instantiates a ground-truth function $F$ by including a column indicating, for each defendant, whether or not they re-offended after two years. Finally, I assume a uniform probability distribution $\mathbbm{P}$ over all input vectors.\par

Equipped with all the inputs for an accuracy measure, I obtain estimates of both the Brier measure (-.459) and the logarithmic measure (-.740) of the accuracy of the COMPAS algorithm on the data set of defendants Angwin et al.\ produced as part of their audit. To obtain these estimates, I simply calculate Equation~\ref{eq:estacc} for the inputs defined above, alternately using the Brier and logarithmic scoring rules to calculate overall accuracy. Next, I split the data set of patients into those who identified as Black and those who are not identified as Black, before calculating the value of \textsc{EqOpp} for this partition of the data set. This value is -.145, with the overall false negative rate for Black defendants being nearly 44\%, as compared to 29\% for non-Black defendants. This replicates Angwin et al.'s finding of significant unfairness against Black defendants on the part of COMPAS. Note that in calculating \textsc{EqOpp} for this partition of the data set, I implicitly followed a set of instructions detailing \textit{how} to split the data for the purpose of calculating \textsc{EqOpp}.\par

Thus, we are able to obtain estimates of two accuracy measures and one fairness measure for the COMPAS algorithm. One could, of course, obtain many more accuracy and fairness measures as well, but for the sake of simplicity, I will stick to these three measures. Next, I define utility functions on all three measures. Note that each measure is maximized at zero, and is otherwise negatively valued. To translate this preference for approaching zero from above, I define the following utility function on all three measures \textsc{Brier}, \textsc{Log}, and \textsc{EqOpp}:
\begin{equation}\label{eq:utilities}
    u_{\textsc{Brier}}(r) = u_{\textsc{Log}}(r) = u_{\textsc{EqOpp}}(r) = |r|^{-1}.
\end{equation}
We assume for the sake of simplicity that none of the measures \textsc{Brier}, \textsc{Log}, or \textsc{EqOpp} is ever maximized at zero. Equipped with these utility functions, and in light of the result presented in the previous section, and letting $\alpha=0$, we can now state the general form of the function \textsc{Overall} measuring the overall value of the COMPAS algorithm for an agent who cares about accuracy (as measured by both the Brier score and the logarithmic score) and equal opportunity:
\begin{multline}\label{eq:COMPASoverall}
\textsc{Overall}(\mathcal{A},\mathbf{v},\textbf{\textsc{Util}},\mathcal{D},F,\mathbbm{P},\mathcal{E},\mathcal{I}) = w_{\textsc{Brier}}\bar{u}_{\textsc{Brier}}(\mathbf{v}) \\ + \  w_{\textsc{Log}}\bar{u}_{\textsc{Log}}(\mathbf{v}) +  w_{\textsc{EqOpp}}\bar{u}_{\textsc{EqOpp}}(\mathbf{v}).
\end{multline}
Note here that $\textbf{\textsc{Util}}$ is the set $\{u_{\textsc{Brier}}, u_{\textsc{Log}},u_{\textsc{EqOpp}}\}$ of utility functions and $\mathbf{v}$ is the vector $[-.459,-.740,-.145]$, representing the values of the measures \textsc{Brier}, \textsc{Log}, and \textsc{EqOpp}, respectively.\par

\begin{figure}
    \centering
    \includegraphics[width=0.8\linewidth]{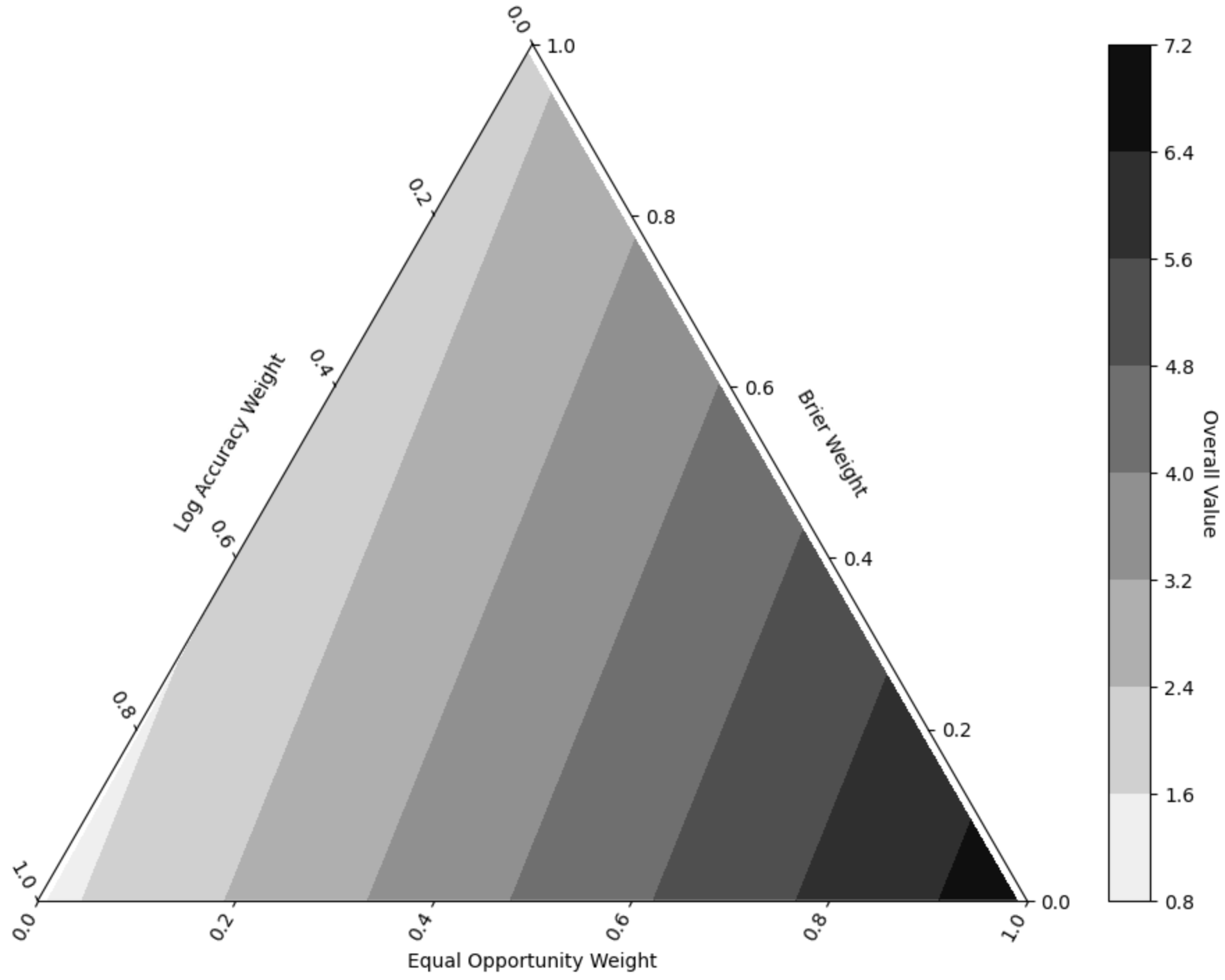}
    \caption{Overall value of the COMPAS predictive algorithm when the weights $w_{\textsc{Brier}}$, $w_{\textsc{Log}}$, and $w_{\textsc{EqOpp}}$ must all be positive and sum to one, and the utility functions $u_{\textsc{Brier}}$, $u_{\textsc{Log}}$, and $u_{\textsc{EqOpp}}$ are defined by Equation~\ref{eq:utilities}.}
    \label{fig:COMPAS1}
\end{figure}

To examine how changes in the weights $w_{\textsc{Brier}}$, $w_{\textsc{Log}}$, and $w_{\textsc{EqOpp}}$ lead to changes in the overall evaluation of the COMPAS algorithm, let us impose the constraints that $w_{\textsc{Brier}}$, $w_{\textsc{Log}}$, and $w_{\textsc{EqOpp}}$ are each positive and must jointly sum to one. Figure~\ref{fig:COMPAS1} shows the overall value of the COMPAS algorithm, as measured by Equation~\ref{eq:COMPASoverall}, for all settings of the weights $w_{\textsc{Brier}}$, $w_{\textsc{Log}}$, and $w_{\textsc{EqOpp}}$ allowed under these constraints. As one can see from this figure, COMPAS is deemed to have greatest overall value when all weight is placed on satisfying equal opportunity. This is unsurprising, given that the value of $u_{\textsc{EqOpp}}$ (6.90) is greater than that of $u_{\textsc{Brier}}$ (2.18) and $u_{\textsc{Log}}$ (1.35).\par

However, while this result is not numerically surprising, it is, perhaps, conceptually surprising. Angwin et al.'s audit of COMPAS is typically thought to have revealed a deep unfairness with respect to the predictions that COMPAS made for Black defendants. So, it is strange that, on this analysis, an agent who cares \textit{only} about equal opportunity would be the one to evaluate COMPAS most positively. There are two ways out of this conundrum. One is to bite the bullet and say that however unfair COMPAS may be, our analysis reveals that COMPAS is \textit{even more inaccurate than it is unfair}, and so it is ultimately plausible that COMPAS would be evaluated most highly by someone who cares solely about equal opportunity.\par

\begin{figure}
    \centering
    \includegraphics[width=0.8\linewidth]{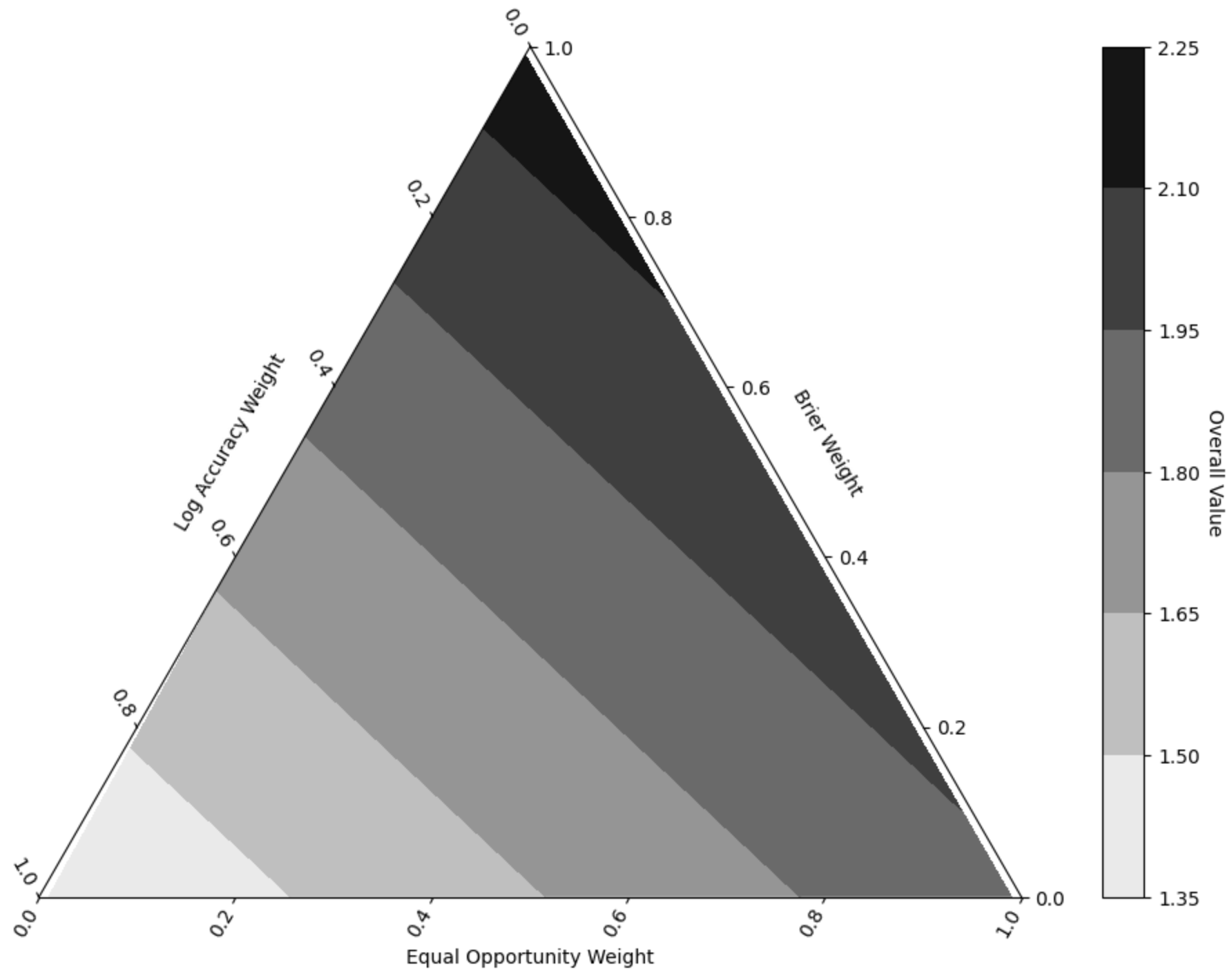}
    \caption{Overall value of the COMPAS predictive algorithm when the weights $w_{\textsc{Brier}}$, $w_{\textsc{Log}}$, and $w_{\textsc{EqOpp}}$ must all be positive and sum to one, the utility functions $u_{\textsc{Brier}}$ and $u_{\textsc{Log}}$ are defined by Equation~\ref{eq:utilities}, and the utility function $u_{\textsc{Brier}}$ is defined by Equation\ref{eq:lastone}.}
    \label{fig:COMPAS2}
\end{figure}

A second approach is to revise the utility function used to evaluate the result of the measure $\textsc{EqOpp}$. We might think that while some amount of inaccuracy, on any measure, is tolerable, less-than-optimal values for a fairness measure like $\textsc{EqOpp}$ must be strongly punished by one's utility function. To that end, let us leave $u_{\textsc{Brier}}$ and $u_{\textsc{Log}}$ unchanged, but let $u_{\textsc{EqOpp}}$ be defined as follows:
\begin{equation}\label{eq:lastone}
    u_{\textsc{EqOpp}}(r) = \ln\left(|r|^{-1}\right).
\end{equation}
On this revised definition, the value of $u_{\textsc{EqOpp}}$ for COMPAS is 1.93, as compared to 2.18 for $u_{\textsc{Brier}}$. Thus, as shown in Figure~\ref{fig:COMPAS2}, according to these utility functions an agent will evaluate COMPAS most favorably if they assign all weight to the Brier measure of accuracy. Nevertheless, there remain some interesting wrinkles:\ in general, assignments of values to weights that regard the Brier measure of accuracy and the equal opportunity notion of fairness as being of similar importance will lead one to regard the COMPAS algorithm more favorably than one who is concerned solely with the logarithmic measure of accuracy. Ultimately, it is up to each agent to decide wether such a recommendation is plausible, or whether the weights and utility functions used to calculate the overall value of COMPAS need to be further adjusted to faithfully recover our intuitions.\par

As a reminder, the preceeding analysis assumes that the desideratum \textbf{D} is an appropriate normative constraint on any aggregate measure of the value of a recidivism prediction algorithm. That is, I have assumed that if our preferences for recidivism prediction algorithms that are Brier-accurate, log-accurate, and satisfy equal opportunity fairness can each be represented by one of three different utility functions, and those utility functions all have equal expectation according to two different probability distributions, then our overall measure of the value of a recidivism prediction algorithm must also have equal expectation according to both probability distributions. For the reasons given above, I take this to be a compelling normative constraint on any aggregate measure of the overall value of a recidivism prediction algorithm.\par

\section{Conclusion}\label{sec:conc}
My aim in this paper has been threefold. First, I aimed to provide a formally rigorous treatment of the trade-off between accuracy and fairness that is inherent in much work on fair prediction. Second, I aimed to use Harsanyi's theorem to show that any such trade-off must take the form of a linear combination of utilities that represent our preferences as to the values of accuracy and fairness measures. Finally, I aimed to use the example of the COMPAS algorithm to demonstrate how these linear combinations can be calculated in practice, and how our choice of utility functions and weights ultimately affects our evaluations of predictive algorithms. I take these initial results to reveal some of the subtle nuances of inferring an agent's overall preferences over prediction algorithms based on their preferences for satisfying competing notions of accuracy and fairness. In future work, I hope to collect empirical data on human judgments of the relative importance of accuracy and fairness in algorithmic decision-making, so as to measure the extent to which actual human judgments can be represented by the model proposed in this paper.\par 

\begin{acks}
   I am grateful to David Watson, Chad Lee-Stronach, Lara Buchak, audiences at Arizona State University, the University of Cincinnati, the University of Kentucky, and Boston University, three anonymous FAccT reviewers and one FAccT AC for their feedback on earlier drafts of this paper. 
\end{acks}

\bibliographystyle{ACM-Reference-Format}
\bibliography{sample-base}
\end{document}